\documentclass[runningheads]{llncs}
\usepackage[T1]{fontenc}
\usepackage{graphicx}
\usepackage{appendix}
\usepackage{algorithm}
\usepackage{amsmath}
\usepackage{algpseudocode}
\usepackage{blindtext}
\usepackage{subcaption}

\usepackage{hyperref}
\usepackage{color}

\usepackage{caption}

\begin{document}
\title{Determining Fetal Orientations From Blind Sweep Ultrasound Video}
\titlerunning{Determining Fetal Orientations From Blind Sweep Ultrasound Video}
%
%

\author{Jakub Maciej Wiśniewski\inst{1} \and
Anders Nymark Christensen\inst{1}\orcidID{0000-0002-3668-3128} \and
Mary Le Ngo\inst{2,3,4} \and
Martin Grønnebæk Tolsgaard\inst{2,4}\orcidID{0000-0001-9197-5564} \and
Chun Kit Wong\inst{1,5}\orcidID{0000-0001-5528-9727}
}  
\authorrunning{J. M. Wiśniewski et al.}
\institute{Technical University of Denmark, Kongens Lyngby, Denmark \\
    \email{\{anym,ckwo\}@dtu.dk} \and
    University of Copenhagen, Copenhagen, Denmark \and
    Slagelse Hospital, Slagelse, Denmark \and
    CAMES Rigshospitalet, Copenhagen, Denmark \and
    Pioneer Centre for AI, Copenhagen, Denmark
}

\maketitle              
\begin{abstract}
Cognitive demands of fetal ultrasound examinations pose unique challenges among clinicians. With the goal of providing an assistive tool, we developed an automated pipeline for predicting fetal orientation from ultrasound videos acquired following a simple blind sweep protocol. Leveraging on a pre-trained head detection and segmentation model, this is achieved by first determining the fetal presentation (cephalic or breech) with a template matching approach, followed by the fetal lie (facing left or right) by analyzing the spatial distribution of segmented brain anatomies. Evaluation on a dataset of third-trimester ultrasound scans demonstrated the promising accuracy of our pipeline. This work distinguishes itself by introducing automated fetal lie prediction and by proposing an assistive paradigm that augments sonographer expertise rather than replacing it. Future research will focus on enhancing acquisition efficiency, and exploring real-time clinical integration to improve workflow and support for obstetric clinicians.

\end{abstract}

\section{Introduction}

\begin{figure}[h]
    \centering
    \begin{minipage}[b]{0.59\textwidth}
        \begin{subfigure}[b]{0.48\textwidth}
            \centering
            \includegraphics[width=\textwidth]{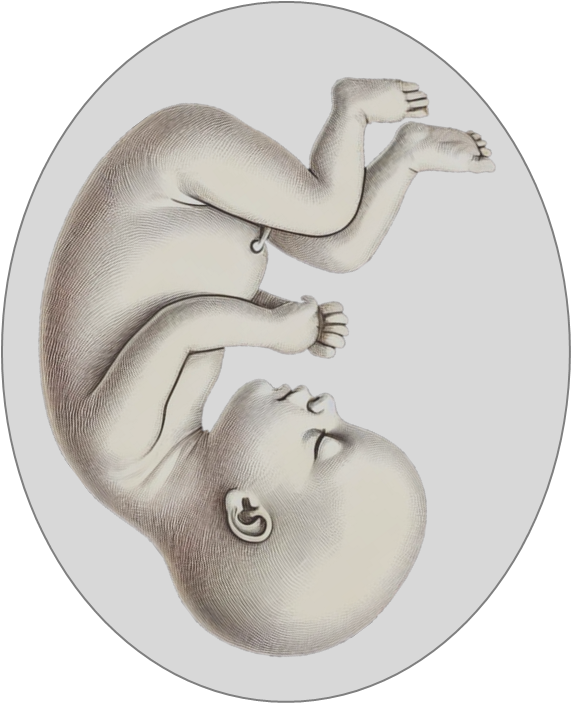}
            \caption{Cephalic, left}
            \label{fig:cephalic_left}
        \end{subfigure}
        \begin{subfigure}[b]{0.48\textwidth}
            \centering
            \includegraphics[width=\textwidth]{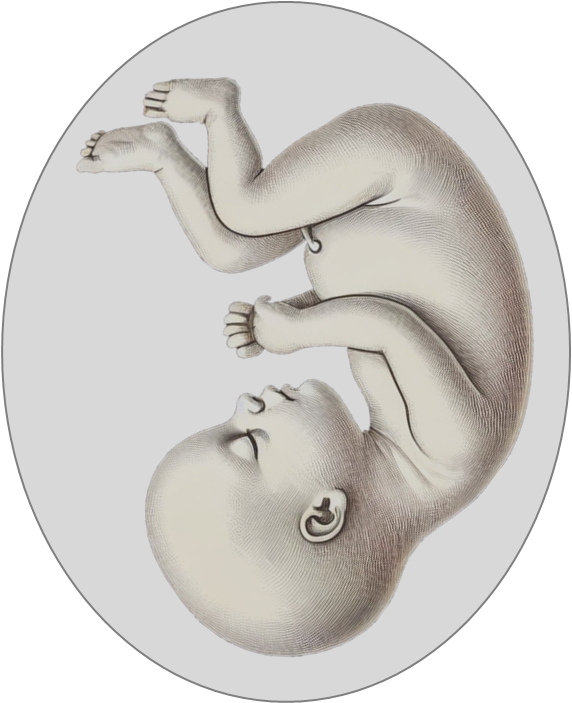}
            \caption{Cephalic, right}
            \label{fig:cephalic_right}
        \end{subfigure}
        
        \begin{subfigure}[b]{0.48\textwidth}
            \centering
            \includegraphics[width=\textwidth]{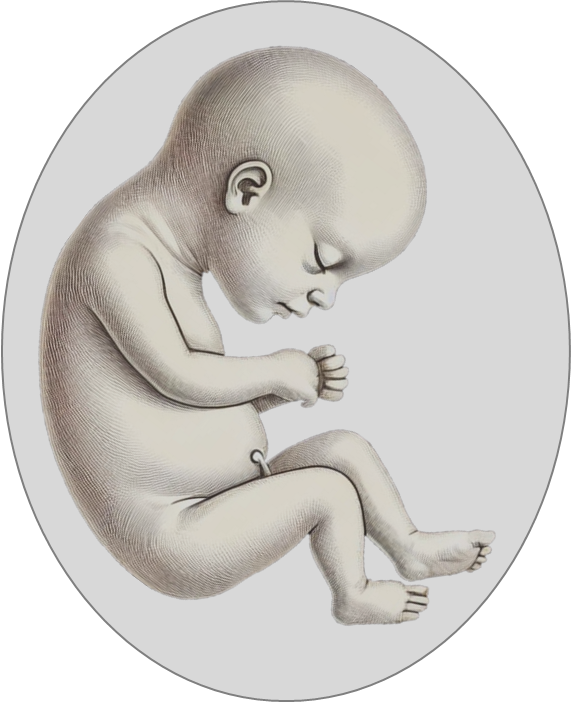}
            \caption{Breech, left}
            \label{fig:breech_left}
        \end{subfigure}
        \begin{subfigure}[b]{0.48\textwidth}
            \centering
            \includegraphics[width=\textwidth]{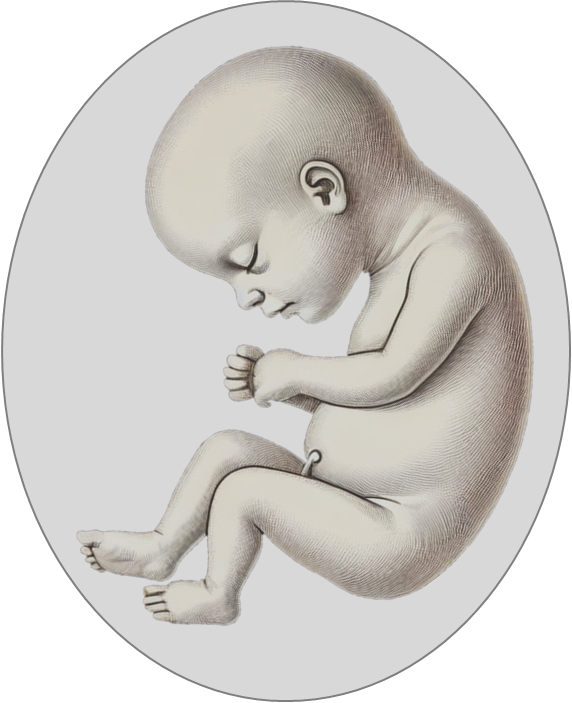}
            \caption{Breech, right}
            \label{fig:breech_right}
        \end{subfigure}
    \end{minipage}
    \begin{subfigure}[b]{0.4\textwidth}
        \centering
        \includegraphics[width=\textwidth]{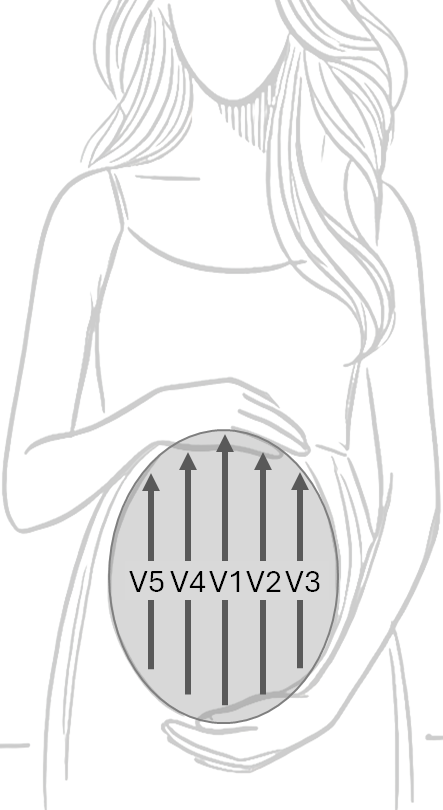}
        \caption{Sweep protocol}
        \label{fig:protocol}
    \end{subfigure}
    \caption{(a-d) The four fetal orientations. (e) Blind sweep protocol across the pregnant abdomen. The sweeps start from the center marked with V1, followed by two on the mother's left (V2, V3) and finally two on the right (V4, V5).}
    \label{fig:fetal_orientations}
\end{figure}

Ultrasound imaging stands as a ubiquitous tool in modern maternal healthcare, playing a crucial role in prenatal monitoring and fetal well-being assessment. During a typical ultrasound examination, sonographers skillfully navigate the probe across the maternal abdomen, mentally constructing a dynamic three-dimensional representation of the fetus~\cite{tolsgaard2017assessment}. This mental model allows them to track fetal position, orientation, and the location of specific anatomical structures, guiding their probe movements to capture the necessary diagnostic information. However, this process demands significant cognitive effort and sustained concentration from the sonographer. High patient volumes, lengthy examination times, and the need for meticulous image interpretation contribute to substantial mental workload~\cite{daugherty2002burnout,johnson2019breaking,singh2017occupational,tran2024incidence} and potential for cognitive fatigue. This is particularly pronounced for sonographers in training or those with less experience~\cite{sharma2021machine}, who may find it challenging to rapidly build and maintain an accurate mental 3D model of the fetus while simultaneously acquiring and interpreting ultrasound images. This cognitive burden can impact workflow efficiency and potentially increase the risk of diagnostic errors.

To address these challenges, numerous prior works have explored automated solutions for analysis of ultrasound examinations~\cite{chen2021artificial,komatsu2021towards,song2024survey}. A significant portion of this research has focused on automating fetal biometry measurements, such as head~\cite{van2019automated,zeng2021fetal} or abdominal~\cite{kim2018machine} circumference and femur length~\cite{hermawati2019automatic}, which is crucial for downstream tasks such as fetal weight estimation and delivery planning. More recent works also explored gestational age estimation from ultrasound videos~\cite{gomes2022mobile,lee2023development,pokaprakarn2022ai}, often acquired via some blind sweep protocols. While these automated systems offer the promise of increased efficiency and reduced operator dependency, they often lack the flexibility required in dynamic clinical settings. Blind sweep protocols, while standardized, may not always capture all necessary information, and sonographer intervention to adjust scanning technique is sometimes necessary. In scenarios where the sonographer deviates from the prescribed protocol to obtain optimal images, these rigid automated systems may fail.

Meanwhile, another significant line of research focuses on fetal standard plane detection~\cite{baumgartner2017sononet,chen2019self,tan2019semi}, which aim to identify and classify key diagnostic planes. Such standard plane detection systems often form an integral part of the fully automated ultrasound analysis systems described previously, or function as retrospective analysis tools. While these standard plane detection systems may also be used to provide real-time assistance, they primarily address the task of identifying specific planes after the sonographer has navigated to them.  They do not directly address the fundamental challenge of helping sonographers in developing the crucial mental 3D model necessary for efficient probe navigation and anatomical localization during the live scan~\cite{wong2024deployment}. The sonographer still needs to mentally keep track of the fetus's overall orientation and position to effectively guide the probe and acquire the necessary standard planes in the first place.

In this work, we propose a different paradigm: an assistive tool designed to alleviate cognitive loading on sonographers during ultrasound examinations. Instead of automated estimation of fetal biometry from blind sweep videos, we focus on determination of coarse fetal orientation – specifically, fetal presentation (cephalic or breech) and lie (facing left or right), with the goal of presenting them as a continuous visual reminder throughout the ultrasound examination (see \autoref{fig:fetal_orientations}). While automated fetal presentation prediction has been explored in prior research~\cite{gleed_statistical_2024}, to the best of our knowledge, this is the first work to specifically address the automated prediction of fetal lie from ultrasound videos. This approach aims to reduce cognitive burden without replacing the sonographer's expertise or limiting their scanning flexibility.

\section{Methods}
\subsection{Blind Sweep Protocol}
\label{sec:blind_sweep_protocol}
For data acquisition, we followed a simple blind sweep protocol inspired by~\cite{pokaprakarn2022ai}, which involved performing five vertical ultrasound sweeps across the pregnant abdomen in the caudal-cranio direction, each traversing from the pubis to the uterine fundus in 10 seconds (see \autoref{fig:protocol}). This ensures comprehensive coverage of the maternal abdomen, and thereby maximizing the probability of capturing the fetal head within the acquired ultrasound videos. 

\subsection{Head Detection and Segmentation}
\label{sec:pcbm_model}
The core of our data processing pipeline relies on a pre-trained progressive concept bottleneck model (PCBM)~\cite{lin_i_2022}, a deep learning model designed for explainable standard plane detection in fetal ultrasound. While PCBM offers a wide array of functionalities, we specifically leverage two of its key capabilities in this work. First, we utilize PCBM's plane classification to identify video frames containing the fetal head within our blind sweep acquisitions. Second, we employ the model's integrated head segmentation module, which provides segmentation masks for two anatomical structures within the head: the thalamus and the cavum septum pellucidum (CSP).

\subsection{Fetal Presentation Classification}
\label{sec:method_presentation_classification}
Inspired by~\cite{gleed_statistical_2024}, we use a simple template matching algorithm to classify the fetal presentation (cephalic versus breech) by making use of temporal information from PCBM's prediction on our blind sweep videos. This is based on the expectation that in cephalic presentations (i.e. head down), the fetal head is encountered earlier in the caudal-cranio sweep, while in breech presentations, the head appears later in the sweep. For each sweep video $i$, we first define a function $f_{v_{i}}(t)$ as the softmax probability output by PCBM for the video frame at temporal point $t$ being classified as containing a fetal head. We then define two exponential template functions, $f_{c}(t)$ and $f_{b}(t)$, to represent the expected temporal distribution of head detections for cephalic and breech presentations:
\begin{align}
    f_{c}(t) &= \frac{\exp(N_{f}-t)}{\exp(N_{f})}, \qquad f_{b}(t) = \frac{\exp(t)}{\exp(N_{f})}
    \label{eq:presentation_pattern_template}
\end{align}
where $N_{f}$ is the total number of frames in the sweep video. These reflect the decreasing likelihood of head detection as the sweep progresses in cephalic cases, and vice versa in breech cases. Finally, we calculate the cosine similarity between the function $f_{v_{i}}(t)$ and each of the template functions, $f_{c}(t)$ and $f_{b}(t)$. The fetal presentation is then classified as cephalic or breech based on which template function yields a higher cosine similarity with $f_{v_{i}}(t)$, followed by majority voting across $i$ as an aggregation step.

\subsection{Fetal Lie Classification}
\label{sec:method_lie_classification}
\begin{figure}[h]
    \begin{subfigure}[b]{0.19\textwidth}
        \centering
        \includegraphics[width=\textwidth]{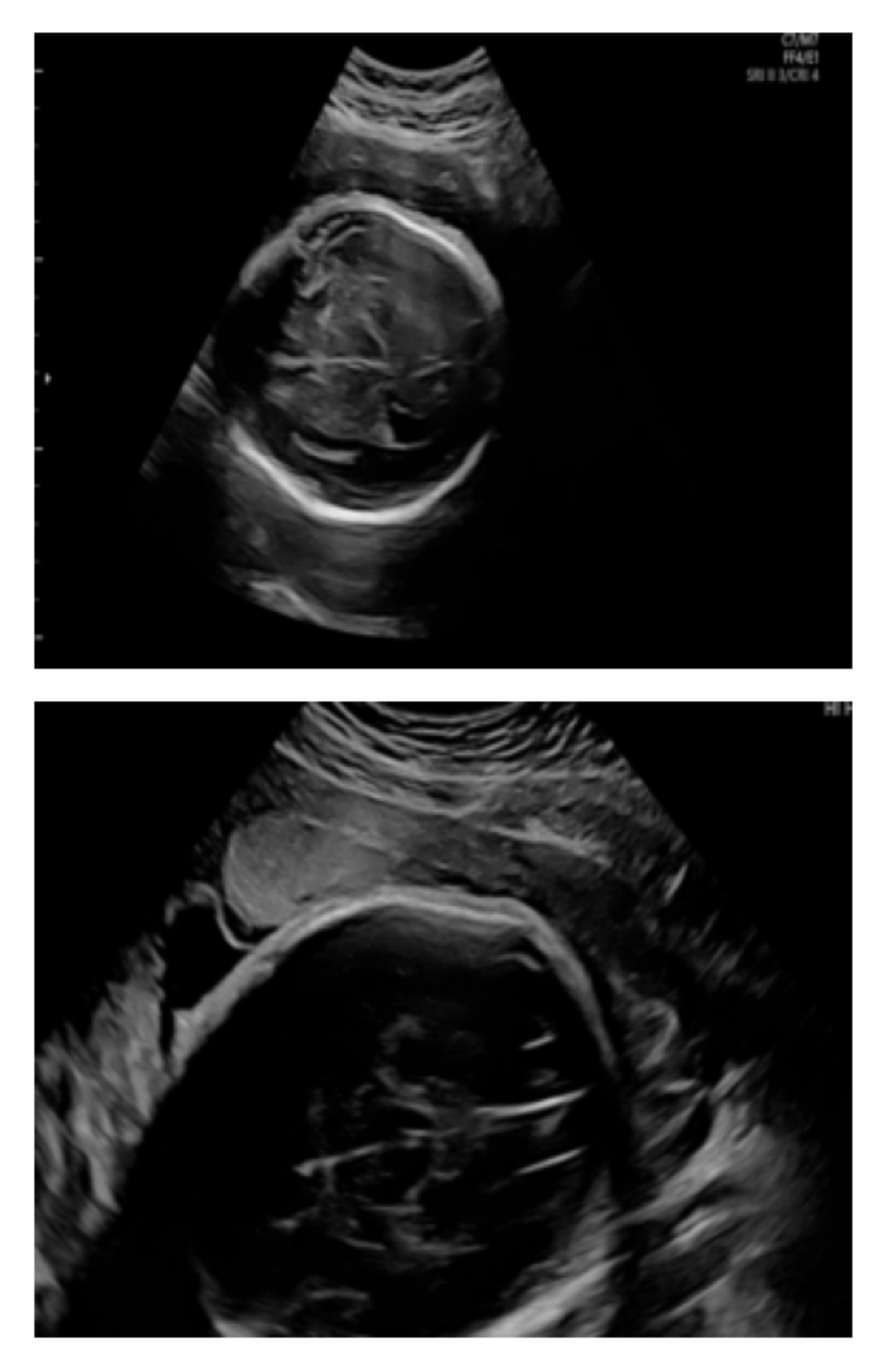}
        \caption*{\scriptsize Detected head}
        \label{fig:flowchart_detected_head}
    \end{subfigure}
    \begin{subfigure}[b]{0.19\textwidth}
        \centering
        \includegraphics[width=\textwidth]{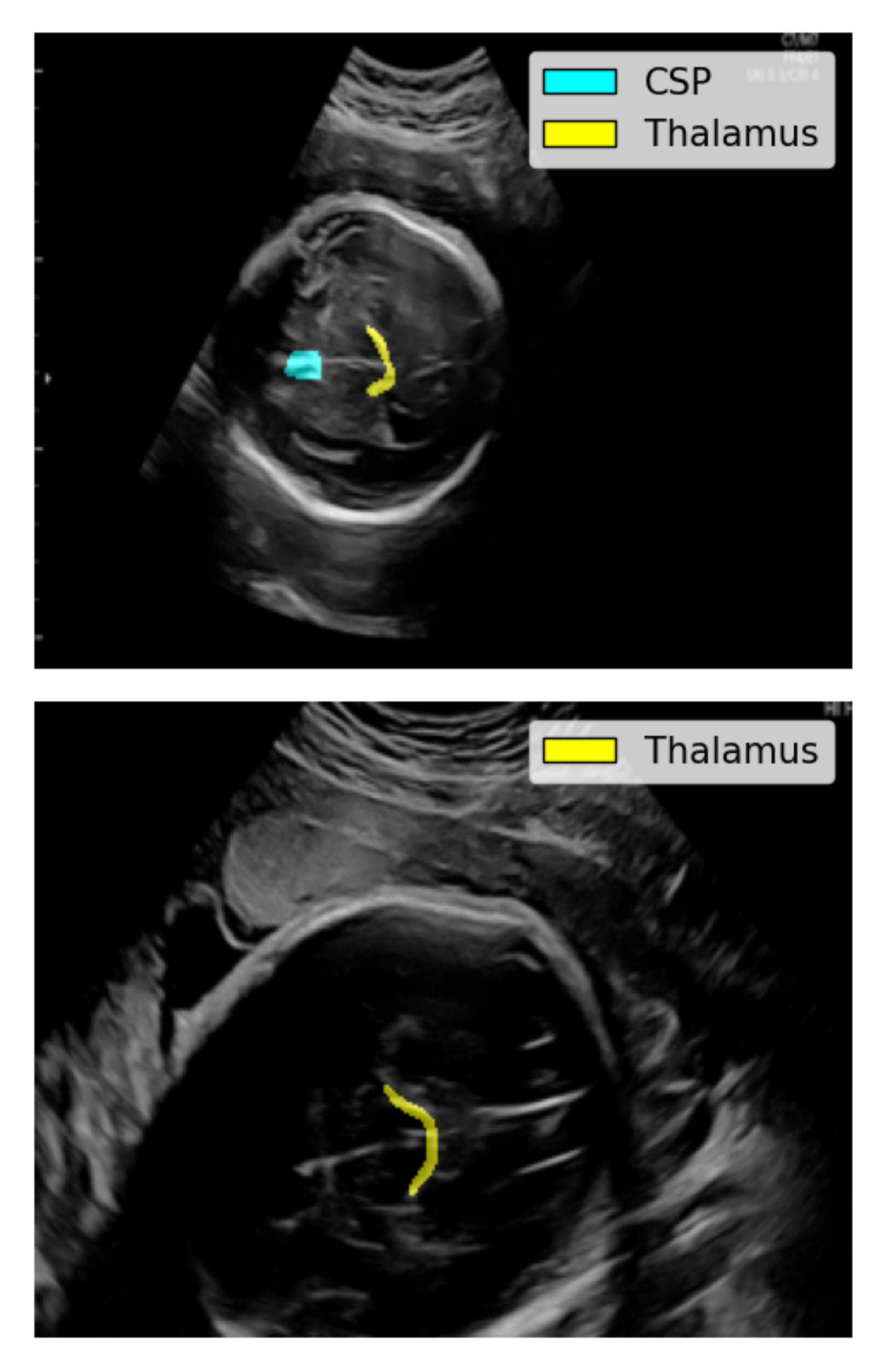}
        \caption*{\scriptsize Segmentation}
        \label{fig:flowchart_segmentation}
    \end{subfigure}
    \begin{subfigure}[b]{0.19\textwidth}
        \centering
        \includegraphics[width=\textwidth]{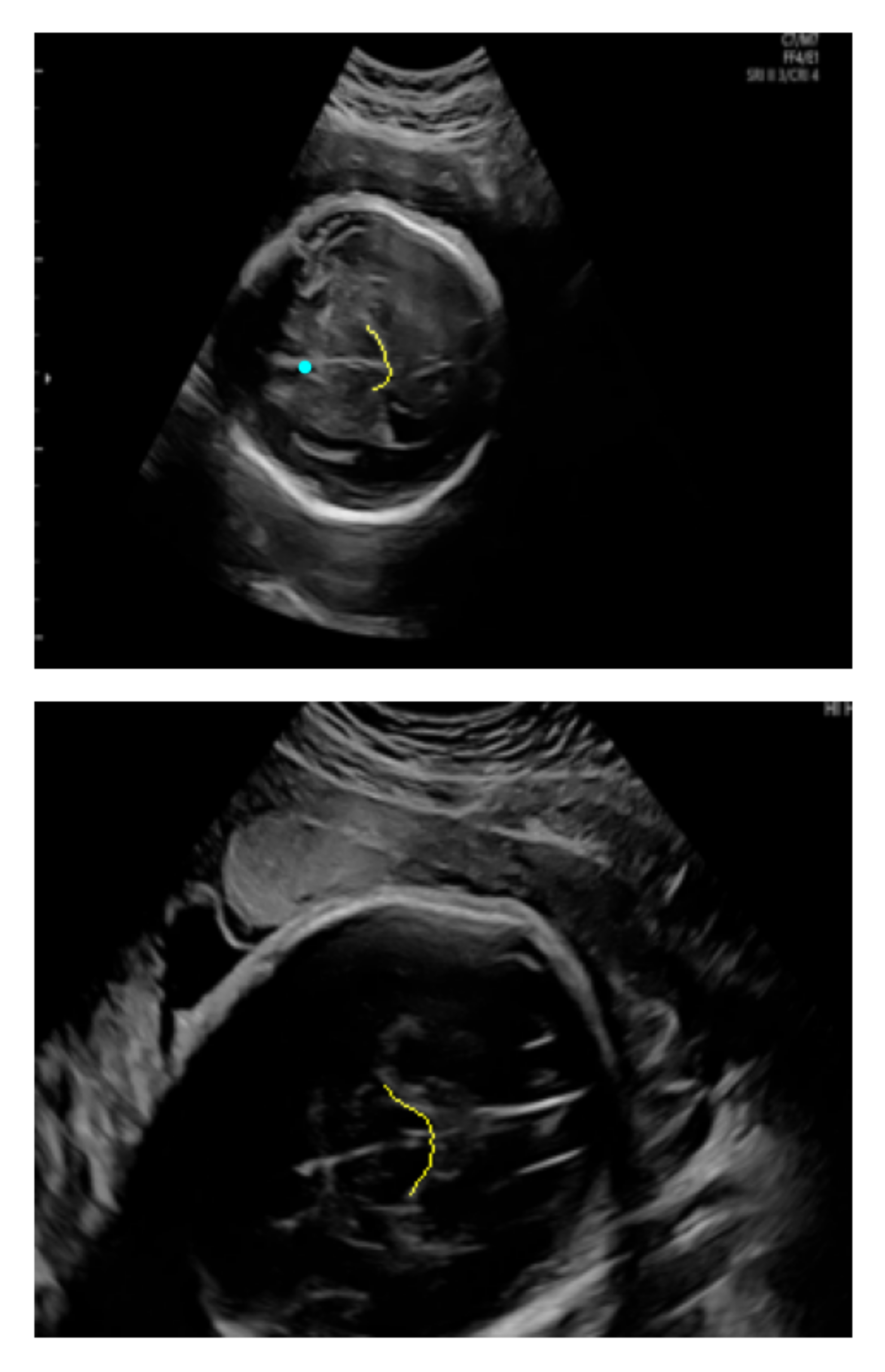}
        \caption*{\scriptsize Skeletonization}
        \label{fig:flowchart_skeletonize}
    \end{subfigure}
    \begin{subfigure}[b]{0.19\textwidth}
        \centering
        \includegraphics[width=\textwidth]{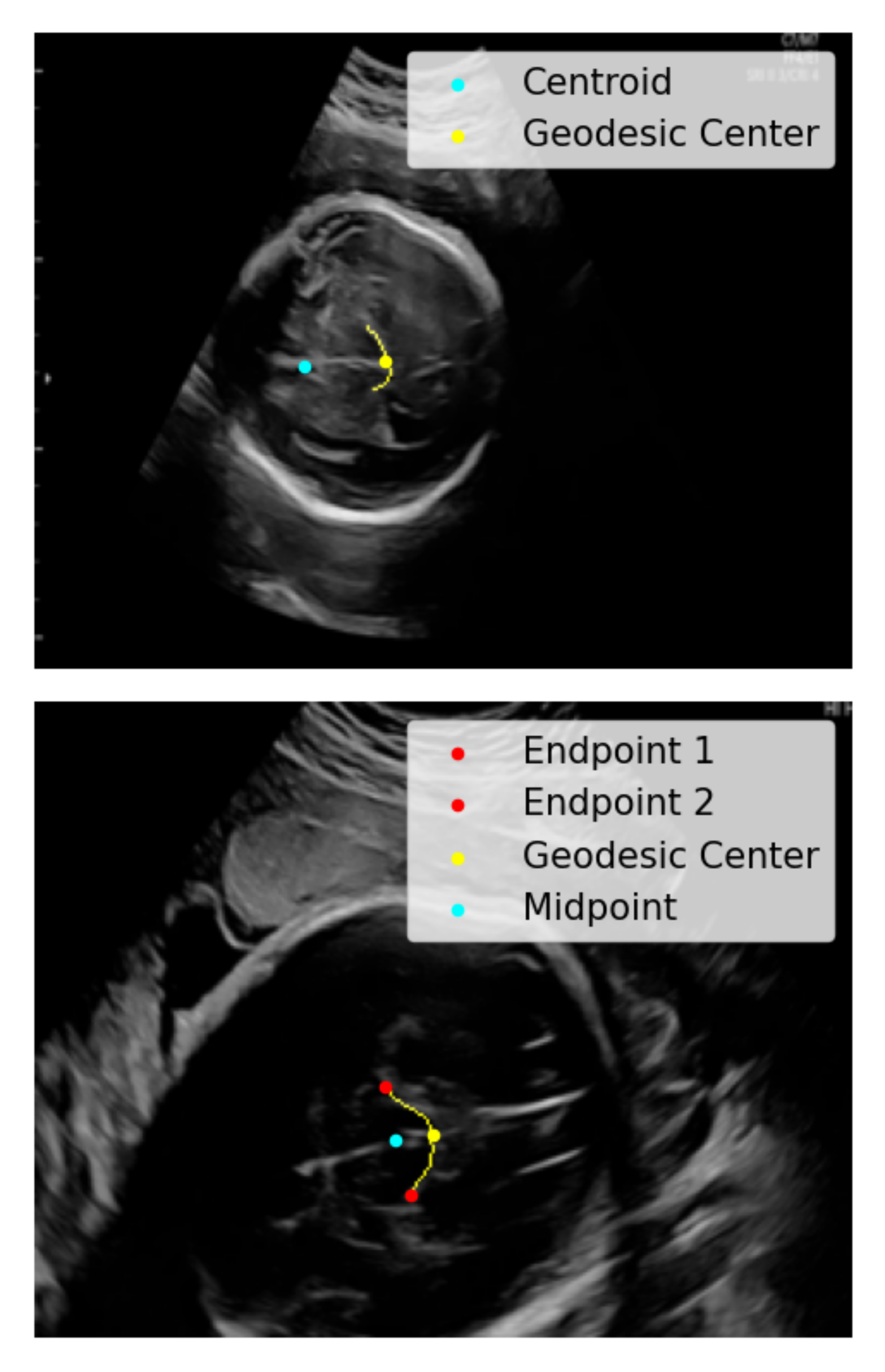}
        \caption*{\scriptsize Extract landmarks}
        \label{fig:flowchart_extracting_landmarks}
    \end{subfigure}
    \begin{subfigure}[b]{0.19\textwidth}
        \centering
        \includegraphics[width=\textwidth]{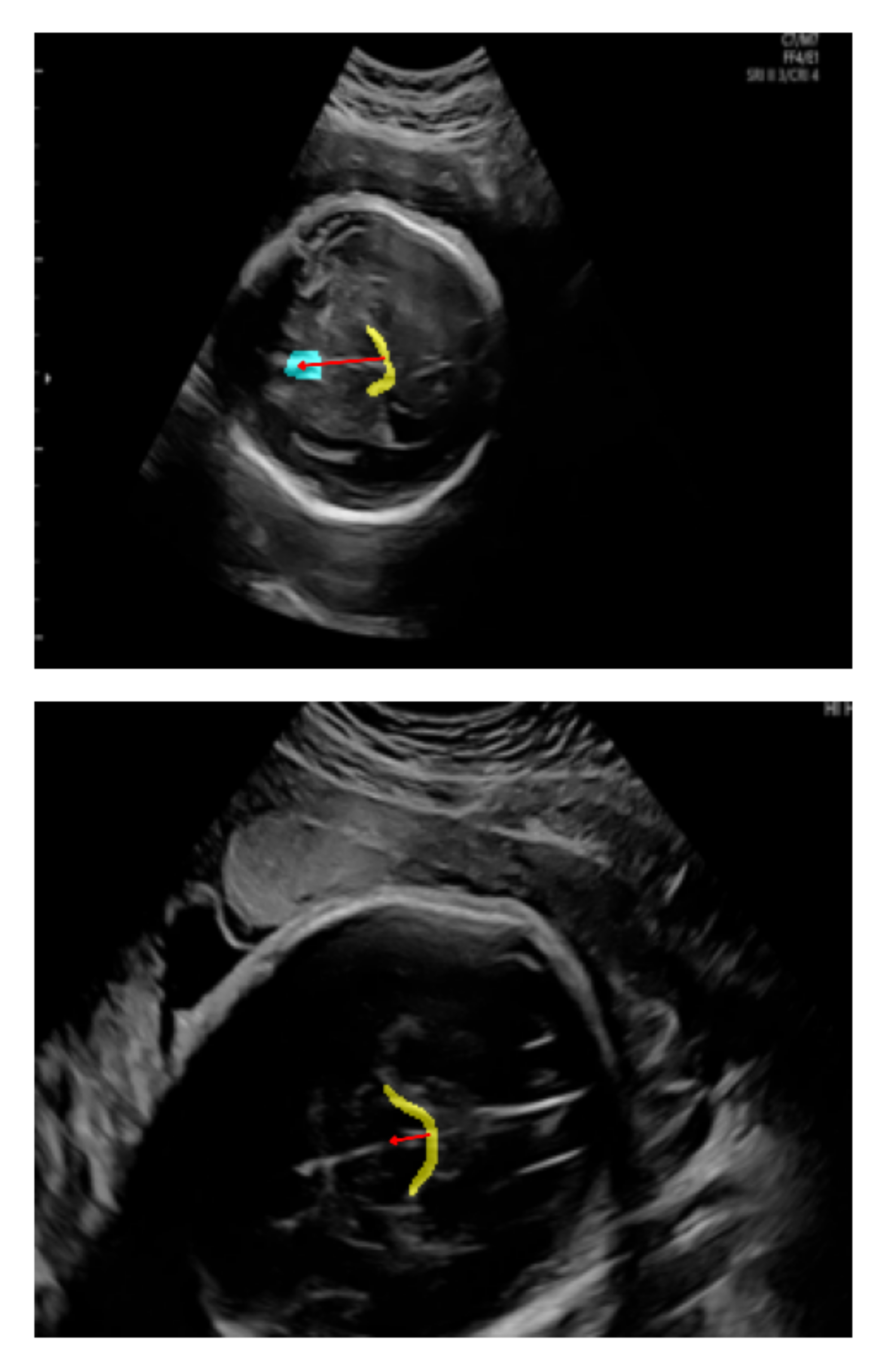}
        \caption*{\scriptsize Get direction}
        \label{fig:flowchart_determine_facing_direction}
    \end{subfigure}
    \caption{The process of estimating fetal lie depicted from left to right. The top row illustrates the main approach of utilizing both thalamus and CSP, while the bottom row illustrates the thalamus-only fallback approach.}
    \label{fig:flowchart}
\end{figure}

For the classification of fetal lying (i.e. facing left versus right), we utilized the segmentation masks of the thalamus and CSP obtained from PCBM, and leveraged the anatomical knowledge that the thalamus is situated posteriorly to the CSP within the fetal head. Consequently, the vector pointing from the thalamus towards the CSP provides an indication of the fetal facing direction. We developed an image processing algorithm to automate this process (see \autoref{fig:flowchart}).

For the thalamus mask, we first determined its geodesic center, which is opted over the centroid. This is because the thalamus exhibits a crecent-shaped morphology, which can lead to the centroid falling outside the actual thalamus structure. On the other hand, the geodesic center, by definition, is guaranteed to lie within the structure. To compute the geodesic center, we first skeletonized the thalamus mask, followed by constructing an adjacency graph of the skeleton pixels. The geodesic center can then be identified as the pixel with the minimum sum of shortest path distances to all other pixels within this graph. For the CSP mask, given its typically blob-like shape, we calculated the centroid. The vector originating from the thalamus geodesic center and terminating at the CSP centroid was then taken as our primary prediction for the fetal facing direction.

To enhance the robustness of our lie determination pipeline, we also incorporated a fallback mechanism to handle scenarios where only the thalamus is reliably segmented, but the CSP is not. In such cases, we again began by skeletonizing the thalamus mask.  Alongside the geodesic center, we also identified the two endpoints of the thalamus skeleton. These endpoints were detected by convolving a kernel function over the skeletonized image.  From the straight line segment connecting these two endpoints, we calculated the orthogonal direction. This orthogonal direction then served as our predicted fetal facing direction when the CSP was not available. However, to ensure the reliability of this thalamus-only approach, we empirically established from early experiments a set of criteria that the thalamus mask must satisfy:
\begin{itemize}
    \item the thalamus mask must consist of a single connected component
    \item it must contain a minimum of 55 pixels
    \item it must exhibit a minimum solidity of 0.82
    \item the distance between the midpoint of the straight line joining the two endpoints and the geodesic center of the thalamus must be at least 1.09 pixels
\end{itemize}
If these criteria were not met, the pipeline would abstain from predicting the lie. 

\section{Experiments and Results}
\label{sec:result}
\subsection{Dataset}
\label{sec:dataset_experiment}
To evaluate the performance of our proposed pipeline, we utilized a dataset acquired with full consent from 13 pregnant subjects in their third trimester. Ultrasound data were collected using our blind sweep protocol (see \autoref{sec:blind_sweep_protocol}), with recordings performed using a research software deployment setup integrated within the clinical ultrasound environment~\cite{wong2024deployment}. Following each blind sweep acquisition, the sonographer who performed the scan also provided the annotation labels immediately following the acquisition. These labels indicated the fetal presentation, categorized as "down" (cephalic) or "up" (breech), and fetal lie, categorized as "left" or "right". In total, 12 cases were annotated as "down" presentation and 1 as "up" presentation, while for fetal lie 5 cases were annotated as "left" and 8 as "right". We acknowledge that this dataset exhibits an imbalance in presentation classes, with a notably smaller number of breech cases. However, this class distribution is reflective of the real-world prevalence of fetal presentations in the third trimester, where cephalic presentation is considerably more common than breech presentation~\cite{fox2006longitudinal}.

\subsection{Fetal Presentation Classification}
\label{sec:result_presentation_classification}
\begin{figure}[t]
  \centering
  \begin{tabular}[b]{c}
    \includegraphics[width=.43\linewidth]{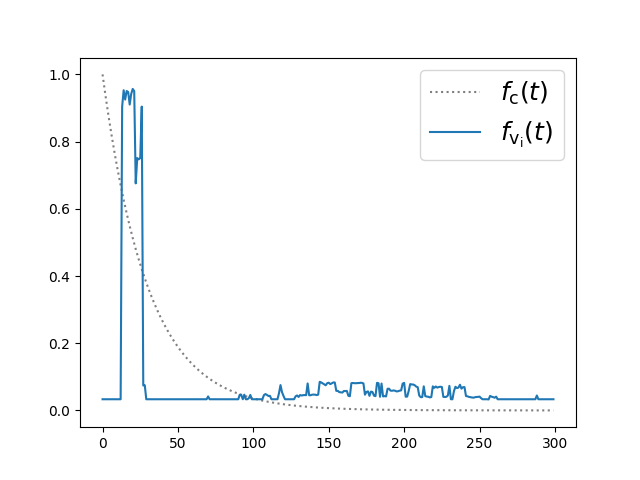} \\
    \small Cephalic
  \end{tabular} \qquad
  \begin{tabular}[b]{c}
    \includegraphics[width=.43\linewidth]{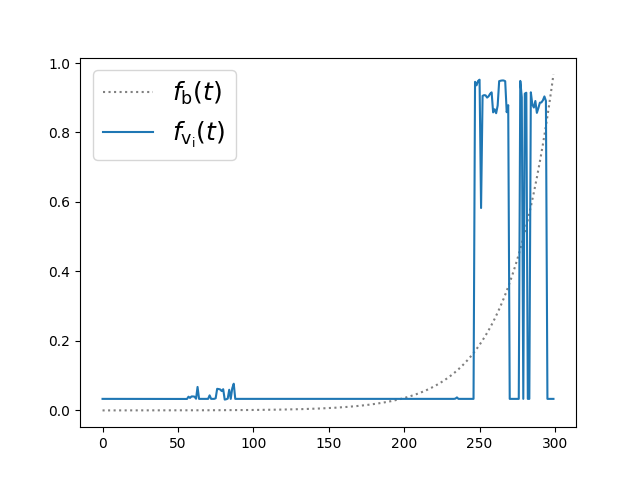} \\
    \small Breech
  \end{tabular}
  \caption{Presentation prediction visualized with sweeps from 2 subjects. The temporal softmax probability for head from PCBM is indicated in blue, while the cephalic and breech template functions (see \autoref{eq:presentation_pattern_template}) are indicated in gray. }
  \label{fig:predictions_up_down}
\end{figure}

We evaluated the performance of our fetal presentation classification method (see \autoref{sec:method_presentation_classification}) on the dataset of 13 subjects. In 2 out of the 13 cases, our pipeline failed to detect the presence of a fetal head within the blind sweep videos, and consequently, presentation classification could not be performed for these cases. For the remaining 11 subjects, where fetal head detection was successful, our template matching algorithm (see \autoref{fig:predictions_up_down}) was able to classify all 10 "down" and the single "up" presentations correctly within this subset of 11 cases.

\subsection{Fetal Lie Classification}
\label{sec:result_lie_classification}

\begin{figure}[h]
    \centering
    \begin{subfigure}[b]{0.3\textwidth}
        \centering
        \includegraphics[width=\textwidth]{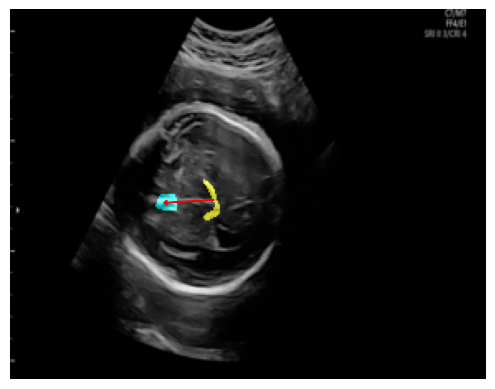}
        \caption{Right}
        \label{fig:patient2_arrow}
    \end{subfigure}
    \begin{subfigure}[b]{0.3\textwidth}
        \centering
        \includegraphics[width=\textwidth]{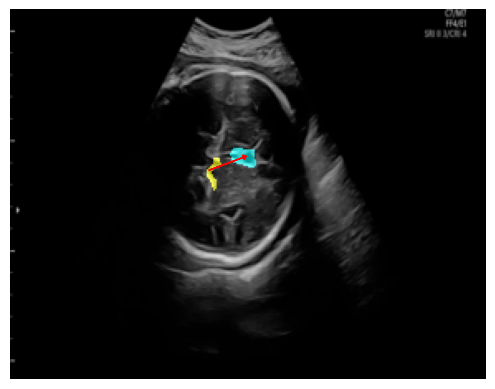}
        \caption{Left}
        \label{fig:patient3_arrow}
    \end{subfigure}
    \begin{subfigure}[b]{0.3\textwidth}
        \centering
        \includegraphics[width=\textwidth]{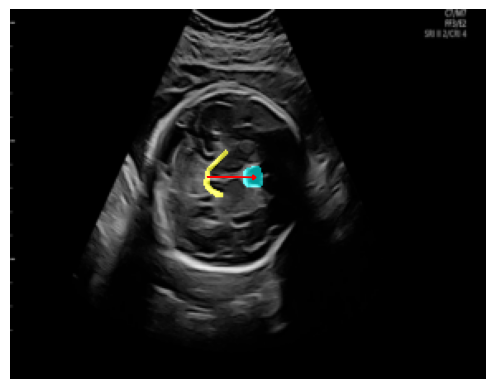}
        \caption{Left}
        \label{fig:patient4_arrow}
    \end{subfigure}

    \begin{subfigure}[b]{0.3\textwidth}
        \centering
        \includegraphics[width=\textwidth]{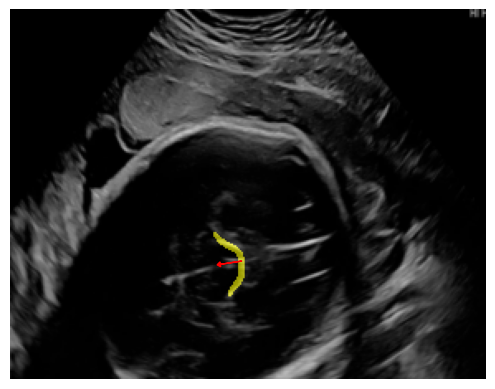}
        \caption{Right}
        \label{fig:patient5_arrow}
    \end{subfigure}
    \begin{subfigure}[b]{0.3\textwidth}
        \centering
        \includegraphics[width=\textwidth]{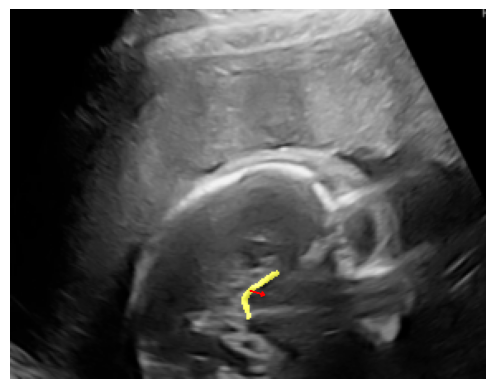}
        \caption{Left}
        \label{fig:patient6_arrow}
    \end{subfigure}
    \begin{subfigure}[b]{0.3\textwidth}
        \centering
        \includegraphics[width=\textwidth]{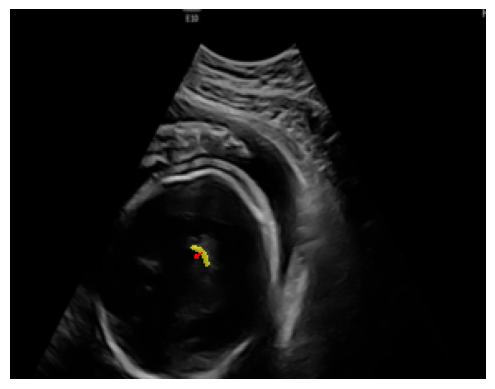}
        \caption{Right}
        \label{fig:patient7_arrow}
    \end{subfigure}

    \begin{subfigure}[b]{0.3\textwidth}
        \centering
        \includegraphics[width=\textwidth]{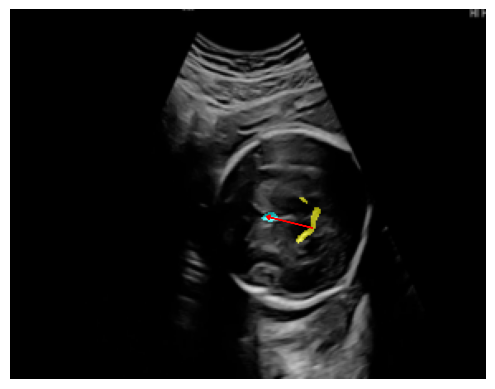}
        \caption{Right}
        \label{fig:patient9_arrow}
    \end{subfigure}
    \begin{subfigure}[b]{0.3\textwidth}
        \centering
        \includegraphics[width=\textwidth]{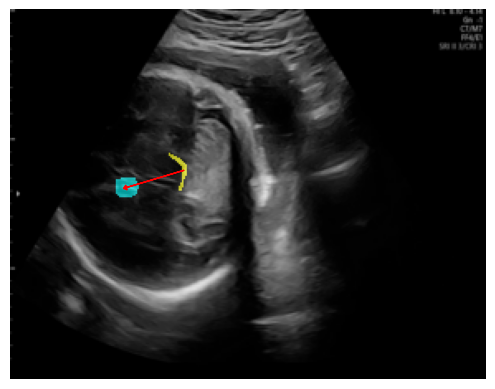}
        \caption{Right}
        \label{fig:patient12_arrow}
    \end{subfigure}
    \begin{subfigure}[b]{0.3\textwidth}
        \centering
        \includegraphics[width=\textwidth]{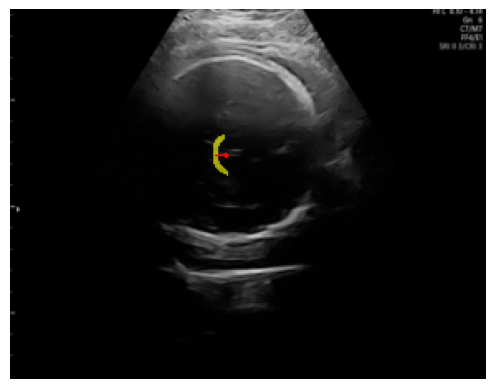}
        \caption{Left}
        \label{fig:patient15_arrow}
    \end{subfigure}
    \caption{Lie prediction visualized for 9 subjects, with ground-truth in the captions.}
    \label{fig:predictions_left_right}
\end{figure}

Next, we evaluated the performance of our fetal lie determination method. Our method, as described in \autoref{sec:method_lie_classification}, predicts a continuous direction vector representing the fetal facing direction. To align with the binary "left" versus "right" lie annotations provided in our dataset, we first discretized our directional predictions by binning them into "left-facing" and "right-facing" categories based on the vector's orientation relative to the image axes.  In addition to the two cases where fetal head detection failed, there were two further cases among the 13 where, despite successful head detection, the PCBM model did not reliably segment the thalamus and CSP, precluding lie determination for those cases.  For the remaining 9 subjects, where our pipeline successfully segmented the necessary anatomical structures and generated a lie prediction, we were able to classify all of them correctly when comparing our binned predictions to the "left" or "right" lie annotations. Among these lie predictions, as visualized in \autoref{fig:predictions_left_right}, some were derived using both thalamus and CSP segmentations, while others relied on the thalamus-only fallback method in cases where CSP segmentation was not reliable.

\section{Discussions}
\subsection{Limitations}
As shown in \autoref{sec:result}, our proposed pipeline relies on the successful detection and segmentation of the fetal head and its internal anatomical structures (thalamus and CSP). While PCBM demonstrates robust performance in our experience, fetal head visualization in ultrasound is not always guaranteed. In later trimesters, the fetal head may descend into the pelvis, becoming partially or fully obscured and thus hindering detection during the blind sweep.  Furthermore, even when the fetal head is detected, the visibility of internal structures like the thalamus and CSP can be compromised by shadowing artifacts or suboptimal image quality, potentially impacting the accuracy of segmentation and subsequent lie determination.  It is important to acknowledge that while our pipeline achieves high accuracy when these steps are successful, its applicability is contingent upon the quality of the acquired ultrasound data.

Another inherent limitation stems from our assumption of limited fetal movement during the blind sweep acquisition.  While fetal movement is a biological reality, it is often negligible for the purposes of coarse presentation and lie determination within the short duration of a sweep.  Fetuses are unlikely to transition from cephalic to breech or left to right lie within a few seconds of scanning. However, we acknowledge that significant or rapid fetal movements during the sweep could potentially introduce inaccuracies in our method.


\subsection{Future works}
One promising avenue for future development is to further simplify our blind sweep protocol to enhance its efficiency and practicality in busy clinical settings. Our current blind sweep protocol, involving five vertical sweeps, was initially designed to maximize the chances of capturing the fetal head regardless of its position within the maternal abdomen. This comprehensive approach aimed to ensure robust head detection and subsequent analysis. However, upon detailed review of our acquired dataset, we observed a recurring pattern: in a majority of successful cases, the fetal head and the anatomical structures necessary for lie determination were adequately visualized within a single vertical sweep, particularly when the sweep was performed along the midline of the abdomen. This suggests that the redundancy provided by five sweeps may not always be necessary for successful presentation and lie assessment, as long as the fetal head can be detected and segmented effectively within that single sweep.

Therefore, the protocol can potentially be simplified to a single vertical sweep together with an adaptive acquisition strategy. In this refined workflow, the system would initially prompt the sonographer to perform only the single midline vertical sweep. The automated pipeline would then analyze the acquired data. If the system encounters challenges – such as failure to detect the fetal head or reliably segment anatomical structures – the system would guide the sonographer to acquire additional vertical sweep lines to improve data quality; otherwise, it would terminate with presentation of the predictions following a successful automated analysis. This adaptive approach aims to strike a balance between minimizing scanning time in typical cases and ensuring reliable results even in more challenging scenarios, ultimately enhancing the clinical utility of our automated pipeline.

\section{Acknowledgment}
This work was supported by the Danish Pioneer Centre for AI (DNRF grant number P1) and SONAI - a Danish Regions’ AI Signature Project.

\bibliographystyle{splncs04}
\bibliography{mybibliography}

\end{document}